\documentclass[letterpaper]{article} 
\usepackage{aaai2027}  
\usepackage[hyphens]{url}  
\usepackage{graphicx} 
\usepackage{natbib}  
\usepackage{caption} 
\usepackage{algorithm}
\usepackage{algorithmic}
\usepackage{amsmath}
\usepackage{amssymb}
\usepackage{siunitx}
\usepackage{multirow}
\usepackage{mathtools}
\usepackage[table]{xcolor}

\definecolor{bestyellow}{HTML}{FFF2CC} 
\definecolor{bestborder}{HTML}{D6B656} 
\definecolor{secondblue}{HTML}{E1F5FE} 
\definecolor{secondborder}{HTML}{0288D1}

\definecolor{gainc}{RGB}{0,128,0}
\definecolor{dropc}{RGB}{178,34,34}
\newcommand{\gain}[2]{$#1_{\textcolor{gainc}{+#2}}$}

\newcommand{\refval}[1]{$#1$}
\newcommand{\best}[1]{%
  \setlength{\fboxsep}{0.8pt}
  \colorbox{bestyellow}{\textbf{#1}}
}
\newcommand{\second}[1]{%
  \setlength{\fboxsep}{0.8pt}%
  \colorbox{secondblue}{\underline{#1}}
}
\usepackage{pifont}
\usepackage{makecell}
\usepackage{array}
\newcolumntype{C}[1]{>{\centering\arraybackslash}p{#1}}

\usepackage{array}

\definecolor{modelrowshade}{RGB}{239,239,239}

\newcolumntype{R}{>{\raggedleft\arraybackslash}p{0.036\textwidth}}
\newcolumntype{Z}{>{\raggedleft\arraybackslash}p{0.040\textwidth}}
\usepackage{newfloat}
\usepackage{listings}
\DeclareCaptionStyle{ruled}{labelfont=normalfont,labelsep=colon,strut=off} 
\floatstyle{ruled}
\newfloat{listing}{tb}{lst}{}
\floatname{listing}{Listing}

\usepackage{booktabs}

\title{From Scoring to Acting: Outcome-Verified Comparative Self-Distillation for LLM Agents}
\author{
    Written by AAAI Press Staff\textsuperscript{\rm 1}\thanks{With help from the AAAI Publications Committee.}\\
    AAAI Style Contributions by Peter Patel Schneider,
    Sunil Issar,\\
    J. Scott Penberthy,
    George Ferguson,
    Hans Guesgen,
    Francisco Cruz\equalcontrib\corresponding,
    Marc Pujol-Gonzalez\equalcontrib\corresponding
}
\affiliations{
    \textsuperscript{\rm 1}Association for the Advancement of Artificial Intelligence\\

    1101 Pennsylvania Ave, NW Suite 300\\
    Washington, DC 20004 USA\\
    proceedings-questions@aaai.org
}

\author{
    Min Yang\textsuperscript{\rm 1,2},
    Jinghua Piao\textsuperscript{\rm 2,3}\corresponding,
    Xu Xia\textsuperscript{\rm 2,4},
    Xiaochong Lan\textsuperscript{\rm 3},
    Jiaju Chen\textsuperscript{\rm 2,5},\\
    Yongshun Gong\textsuperscript{\rm 1},
    Yong Li\textsuperscript{\rm 2,3}\corresponding
}
\affiliations{
    \textsuperscript{\rm 1}Shandong University\quad
    \textsuperscript{\rm 2}Zhongguancun Academy\quad
    \textsuperscript{\rm 3}Tsinghua University\\
    \textsuperscript{\rm 4}Southeast University\quad
    \textsuperscript{\rm 5}University of Science and Technology of China\\
    minyang@mail.sdu.edu.cn\quad
    \{pjh22, lanxc22\}@mail.tsinghua.edu.cn\quad
    s-xx25@bza.edu.cn\\
    cjj01@mail.ustc.edu.cn\quad
    ysgong@sdu.edu.cn\quad
    liyong07@tsinghua.edu.cn
}

\title{From Scoring to Acting: Outcome-Verified Comparative Self-Distillation for LLM Agents}
\author{
    Xu Xia\textsuperscript{\rm 1,2},
    Jinghua Piao\textsuperscript{\rm 2,3}\corresponding,
    Min Yang\textsuperscript{\rm 2,4},
    Xiaochong Lan\textsuperscript{\rm 3},
    Jiaju Chen\textsuperscript{\rm 2,5},\\
    Yong Li\textsuperscript{\rm 2,3}\corresponding
}
\affiliations{
    \textsuperscript{\rm 1}Southeast University\quad
    \textsuperscript{\rm 2}Zhongguancun Academy\quad
    \textsuperscript{\rm 3}Tsinghua University\\
    \textsuperscript{\rm 4}Shandong University\quad
    \textsuperscript{\rm 5}University of Science and Technology of China\\
    s-xx25g@bza.edu.cn\quad
    \{pjh22, lanxc22\}@mail.tsinghua.edu.cn\quad
    minyang@mail.sdu.edu.cn\quad
    cjj01@mail.ustc.edu.cn\quad
    ysgong@sdu.edu.cn\quad
    liyong07@tsinghua.edu.cn
}

\begin{document}

\maketitle

\begin{abstract}
Recent work on LLM agents is shifting from external capability elicitation to capability internalization, enabling agents to retain useful skills without retrieval at inference time. On-policy self-distillation (OPSD) offers a promising direction, but many existing methods typically supervise students by scoring actions along student-generated trajectories. Such supervision has two limitations: teacher preferences are not validated by environment outcomes, and action-level scores underuse information from student rollouts, teacher rollouts, and their behavioral relationship. We therefore advocate outcome-verified teacher supervision and comparative learning over teacher–student trajectories. Based on this view, we propose Outcome-Verified Comparative Self-Distillation (OVCSD). OVCSD organizes failed student rollouts into a prefix tree, adaptively invokes a skill-conditioned teacher from student-reached states, and retains only outcome-verified successful continuations. It then applies localized comparative learning at the first state-aligned divergence and distills the post-divergence teacher suffix to transfer completion behavior. Experiments on ALFWorld and WebShop across three model scales show that OVCSD consistently outperforms skill-free RL and existing self-distillation baselines, achieving up to 29.7 and 5.4 absolute success-rate gains over the strongest baselines on ALFWorld and WebShop, respectively, while adding less than 3\% privileged interaction during training.

\end{abstract}
\begin{links}
    \link{Code}{https://github.com/shane990928-xia/OVCSD}

\end{links}

\section{Introduction}
Large language model agents advance through two complementary paths: scaling pretraining \cite{kaplan2020scaling,hoffmann2022training}, and acquiring reusable capabilities via richer forms of supervision during post-training \cite{yao2022react,wang2026rlanything,ppo,grpo,wei2026agentic,yu2026dapo}. External information, such as skills \cite{skillrl,nie2026skillgraph}, memories \cite{hu2025memory,ouyang2025reasoningbank,yan2026memory,zhou2025mem1}, and expert demonstrations ~\cite{zeng2024agenttuning,xiao2025limi}, can unlock capabilities that remain beyond an agent's standalone repertoire.  However, these gains often reflect temporary capability elicitation rather than lasting acquisition: elicited behaviors fade as soon as the external information is withdrawn. This gap has spurred growing interest in capability internalization, which seeks to consolidate externally elicited behaviors into model parameters, enabling the agent to deploy those capabilities directly and independently
 \cite{skill0,shi2026skill1,skillsd,sdar,tu2026ucob,huang2026skill}. 
\begin{figure*}[t]
\centering
\includegraphics[width=0.95\textwidth]{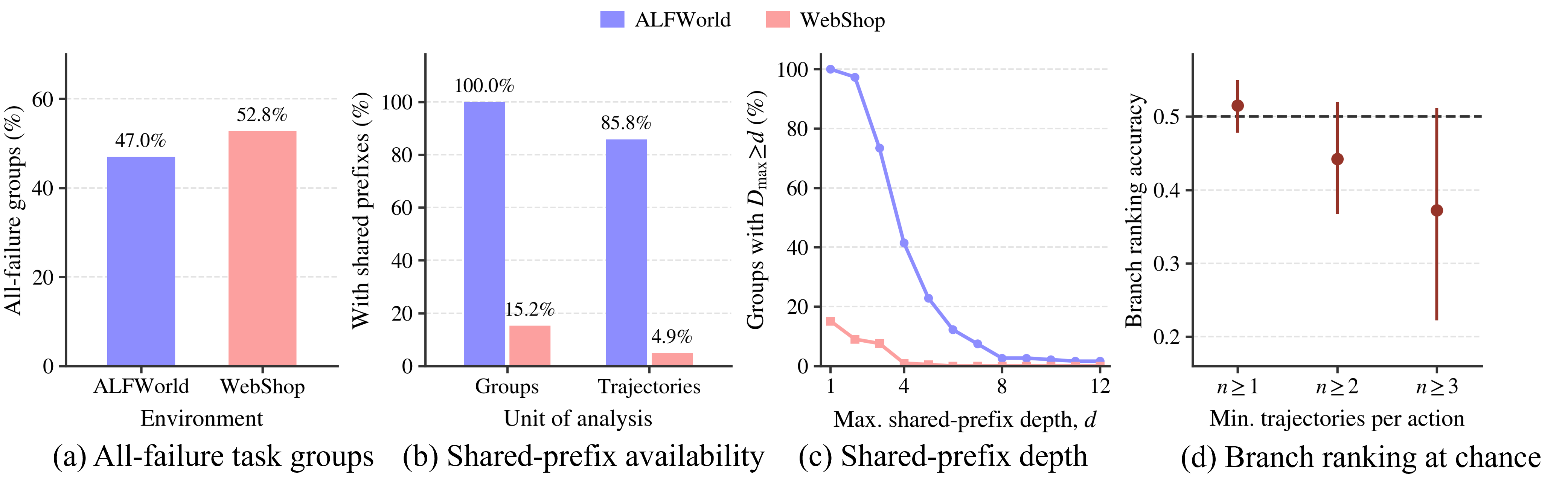}
\caption{%
Preliminary observations on ALFWorld and WebShop.
(a) Frequency of all-failure groups.
(b) Shared-prefix coverage within these groups.
(c) Survival curve of maximum shared-prefix depth $D_{\max}$.
(d) Accuracy of skill-conditioned likelihood ranking at the first divergence
($0.515$, $0.443$, $0.372$ for minimum action support $n=1,2,3$; dashed line
denotes chance).%
}
\label{fig:observations}
\end{figure*}
A straightforward approach is supervised fine-tuning (SFT) on successful privileged trajectories \cite{ouyang2022training,zelikman2022star}. However, collecting such demonstrations is expensive, and treating every action may internalize irrelevant behavior alongside effective decisions \cite{chen2025atlas,lyustudent}. Additionally, SFT relies on offline data distributions, making policies susceptible to distribution shifts during online interaction \cite{lauffer2025imitation,chu2025sft}. To address these limitations, on-policy self-distillation(OPSD) \cite{opsd,agarwal2024policy} collects trajectories from student policy and uses the same model with extra information as the teacher to provide stronger supervision on student-visited states. However, existing methods often use the teacher to score actions taken along student trajectories \cite{li2026policy}. Such methods are based on an important assumption: the teacher’s local preference for an action is a reliable proxy for the action’s contribution to eventual task success \cite{fu2026revisiting,tang2026rewarding,wang2026not}. 

This assumption is particularly fragile in long-horizon agent tasks. A highly scored action may still lead to failure because of subsequent decisions, whereas a low score tells the student what to avoid without showing what to do instead. Addressing these limitations requires both concrete teacher actions and environment-verified outcomes. 
Obtaining such teacher trajectories raises a second question: how should they be used for learning? Existing methods leave much of their information underexploited in three respects. First, failed student rollouts are often reduced to terminal rewards or discarded altogether, despite revealing where the student fails and what progress it has already made. Second, teacher trajectories are commonly distilled uniformly, without distinguishing the decisions that correct the failure from the subsequent behavior that completes the task. Third, teacher and student trajectories are rarely aligned and directly compared at the states where their behaviors diverge, making it difficult to identify the differences associated with recovery. Effective learning from teacher trajectories therefore requires fully exploiting three complementary sources of information: the student’s own rollouts, the teacher’s outcome-verified behavior, and the relationship between them.

We conduct a fine-grained analysis of these three information sources and uncover three observations that directly inform the design of an on-policy self-distillation method. \textbf{First,} failed rollouts are outcome-equivalent but behaviorally structured. Even when every rollout receives the same failure reward, multiple trajectories can share intermediate states and valid prefixes. These shared states expose reusable student progress that is invisible to group-relative terminal rewards. \textbf{Second,} the usefulness of privileged intervention is both state- and outcome-dependent. Shared states occur at different depths across tasks, so a fixed intervention point either wastes valid student progress or intervenes too late. Moreover, the teacher's local action preference is an unreliable proxy for eventual task success—scores alone do not guarantee good outcomes. A teacher should therefore act from adaptively selected student-reached states, with its continuation accepted only after environment-verified success. \textbf{Third,} verified teacher and student branches exhibit a change in comparability at their first state-aligned divergence. Before this point, their behaviors are identical and provide no decision-level contrast. At the divergence, they face the same state but select different actions; afterward, their histories separate and no longer support step-wise comparison. Local corrective credit and subsequent completion behavior should therefore be learned through different objectives.

Based on these observations, we propose Outcome-Verified Comparative Self-Distillation (OVCSD), an on-policy learning framework that turns student failures and teacher interventions into outcome-grounded comparative supervision. 
To exploit the structure of student failures, OVCSD organizes failed trajectories into a prefix tree and searches for intervention states shared by multiple student branches. It begins from deeper student-reached states to preserve as much valid student progress as possible and to enable comparison under the same environment state. When the teacher cannot recover from a deep state, OVCSD progressively backtracks to earlier states, adapting the intervention scope rather than relying on a fixed rollout position.
To obtain supervision grounded in both action and consequence, the privileged teacher acts from the restored student state and its continuation is accepted only when it completes the task according to the environment reward. OVCSD therefore learns from executable, outcome-verified teacher behavior rather than from teacher scores alone.
Finally, OVCSD explicitly compares each successful teacher continuation with the failed student branches it replaces. At their first state-aligned divergence, it increases the probability of the verified teacher action while decreasing the probabilities of the failed student alternatives. After this localized divergence update, it distills the teacher’s remaining suffix into the skill-free policy, teaching the student how to complete the task under the corrected trajectory.
In summary, our main contributions are threefold:
\begin{itemize}
\item \textbf{Outcome-verified self-distillation.}
We introduce OVCSD, a framework that internalizes privileged guidance through
successful environment interaction rather than teacher scores alone.

\item \textbf{Alignment-aware learning.}
We identify state-aligned comparison as the key to learning from successful
teacher behavior and failed student exploration in multi-turn tasks.

\item \textbf{SOTA skill-free performance.}
Across ALFWorld, WebShop, and three model scales, OVCSD consistently achieves
the best skill-free success rate, outperforming the strongest baseline by up
to 29.7 points on ALFWorld and 5.4 points on WebShop.
\end{itemize}

\section{Observations}
\label{sec:observations}

We analyze eight skill-free rollouts for each of 400 tasks from a frozen SFT
model on ALFWorld and WebShop. Panel~(d) additionally scores their actions
under the skill-conditioned view of the same model; no policy update is used.

\paragraph{Observation 1: Failed rollouts retain shared state structure.}
All trajectories fail with identical reward in 47.0\% of ALFWorld groups and
52.8\% of WebShop groups (Figure~\ref{fig:observations}a). Nevertheless,
shared prefixes occur in 100\% and 15.2\% of these groups, covering 85.8\% and
4.9\% of their trajectories, respectively
(Figure~\ref{fig:observations}b). Terminal failure therefore hides reusable
student-reached states, motivating a prefix-tree representation.

\paragraph{Observation 2: Useful intervention is state- and outcome-dependent.}
Shared-prefix depth differs substantially across groups: 97.3\% of ALFWorld
all-failure groups reach depth two, compared with 9.0\% on WebShop
(Figure~\ref{fig:observations}c). Moreover, skill-conditioned likelihood ranks
divergent actions by their empirical outcomes with only 51.5\% accuracy and
remains at or below chance with greater action support
(Figure~\ref{fig:observations}d). OVCSD therefore selects intervention states
adaptively and accepts teacher behavior only after environment-verified
execution.

\paragraph{Observation 3: Comparability changes at the first divergence.}
Before the first state-aligned divergence, successful teacher and failed
student branches are identical and provide no decision-level contrast. At the
divergence, they choose different actions under the same state; afterward,
their histories separate and no longer admit state-wise comparison. This structure motivates a localized comparative objective at the divergence and a separate distillation objective over the remaining teacher suffix.

Together, these observations motivate OVCSD's three components: prefix-tree
construction, adaptive outcome-verified intervention, and alignment-aware
dual-objective distillation.

\begin{figure*}[t]
\centering
\includegraphics[width=0.9 \textwidth]{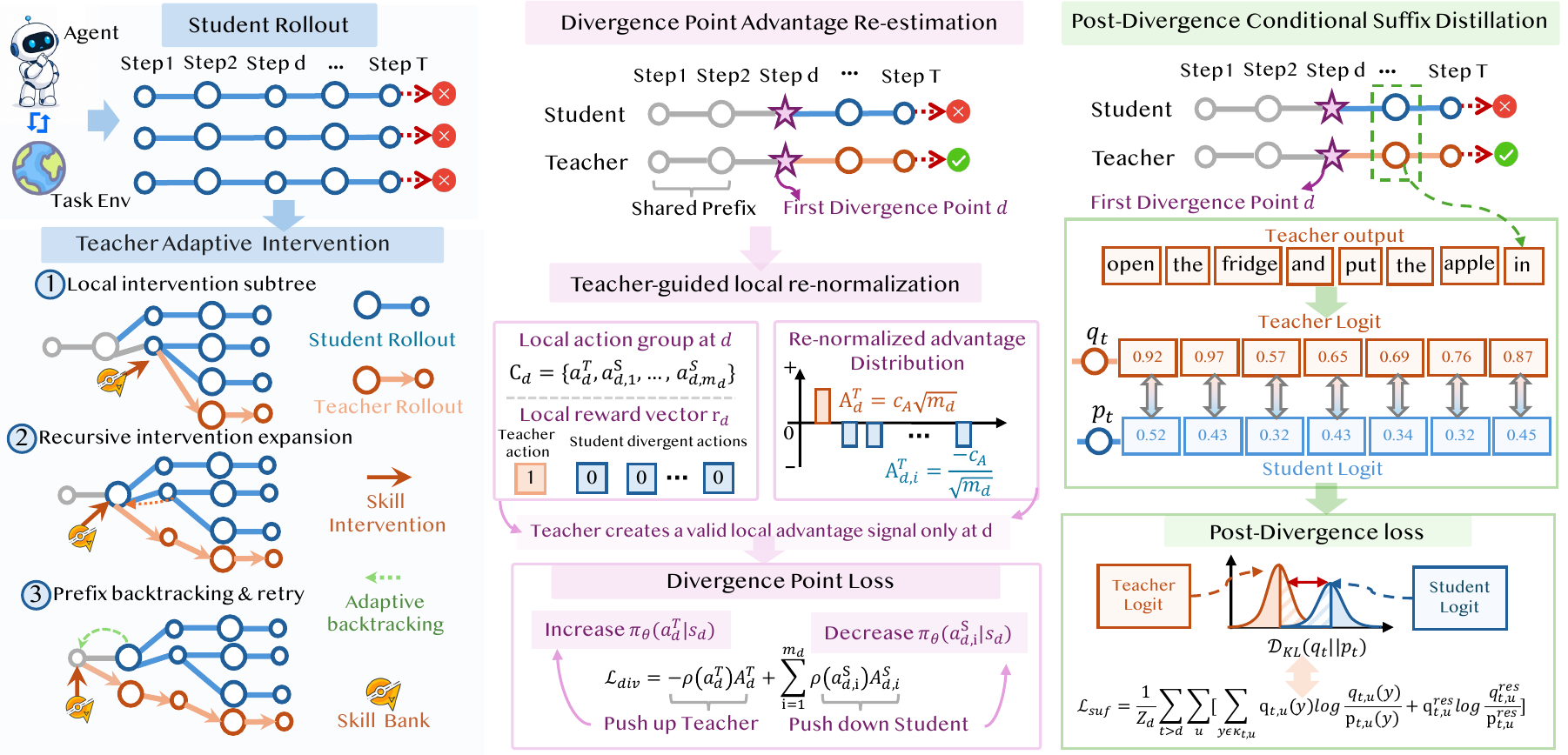}
\caption{Overview of OVCSD. Skill-free student rollouts provide the base group-relative update. For eligible all-failure groups, OVCSD performs state-aligned teacher intervention, constructs local comparative advantages at verified first aligned divergences, and distills the corresponding post-divergence suffixes. Inference uses the skill-free policy alone.}
\label{fig:training-pipeline}
\end{figure*}

\section{Related Work}

\paragraph{Agent Reinforcement Learning and Skill Internalization.}
Recent work improves multi-turn LLM agents through reinforcement learning from
student-generated trajectories. Search-R1~\cite{jin2025search} and
GiGPO~\cite{feng2026group} extend RL to search and long-horizon interaction,
while RAGEN~\cite{wang2025ragen} highlights the difficulty of learning from
sparse trajectory-level rewards. Another line augments agents with retrieved
skills. SkillRL~\cite{skillrl} and SkillGraph~\cite{nie2026skillgraph} construct evolving skill libraries, whereas SKILL0~\cite{skill0} gradually removes skill context to encourage skill-free deployment. These methods either rely on student-generated rewards or retain external guidance during part of the learning process.

\paragraph{On-Policy Self-Distillation.}
Skill-SD~\cite{skillsd} and SDAR~\cite{sdar} use a skill-conditioned teacher to
score actions sampled by a skill-free student. However, the teacher does not
interact with the environment, so its local preferences are not verified by
task outcomes. SIRI~\cite{he2026siri} instead distills from executed
skill-augmented rollouts, but launches them from the initial state and selects
supervision using group-level returns. 

A detailed comparison with existing agent RL and self-distillation methods
is provided in supplementary material.

\section{Method}
\label{sec:method}

\subsection{Overview and Base Student Rollout}

Outcome-Verified Comparative Self-Distillation (OVCSD) converts
uninformative student failures into outcome-grounded supervision for a
skill-free policy. Its design follows the three observations established in the preceding section. OVCSD first uses the shared structure of failed
rollouts to identify student-reached intervention states. It then allows a
privileged teacher to act from these states and retains only continuations
whose success is verified by the environment. Finally, it compares each
verified teacher continuation with the failed student branches in two ways:
the first aligned divergence provides localized corrective credit, while the
remaining teacher suffix provides completion supervision. For a task instance $x$, a trajectory is
\begin{equation}
\tau=(o_0,a_0,o_1,a_1,\ldots,o_T),
\end{equation}
with terminal reward $R(\tau)$. At each update, we freeze the skill-free
rollout policy $\pi_{\bar\theta}^{0}$ and sample a group of $K$ trajectories:
\begin{equation}
D_x^0=\{\tau_i^0\sim\pi_{\bar\theta}^{0}\}_{i=1}^{K},
\qquad
\widehat A_i=
\frac{R_i-\bar R}{s_R+\epsilon_{\mathrm{std}}}.
\end{equation}
These trajectories define the standard group-relative reinforcement-learning
objective. OVCSD intervenes only when all trajectories fail and their rewards
provide no meaningful contrast:
\begin{equation}
z_x^{\mathrm{fail}}
=
\mathbb{I}\!\left[\max_i R_i<r_{\mathrm{succ}}\right]
\mathbb{I}\!\left[s_R\leq\epsilon_R\right].
\end{equation}
For such groups, the original trajectory advantages are set to zero and are
replaced only where outcome-verified comparative supervision is available.

\subsection{Adaptive Outcome-Verified Intervention}

\paragraph{Using shared failure structure.}
Observation~1 shows that failed trajectories may share valid intermediate
progress despite receiving identical terminal rewards. OVCSD therefore
organizes the failed trajectories into a prefix tree. Each node $v$ represents
a student-reached shared state, has depth $d_v$, and supports the subset of
trajectories $M(v)$ that follow the corresponding prefix.
OVCSD prioritizes deeper eligible nodes, since they preserve more student
progress. The environment state at each candidate node is restored through
snapshots or prefix replay and checked for consistency. When no non-trivial
shared state is available, OVCSD falls back to a restorable state from an
individual failed trajectory. Implementation details of transition
canonicalization and state restoration are provided in
supplementary material.

\paragraph{Adaptive intervention and outcome verification.}
The teacher is the same frozen policy snapshot equipped with privileged skill
context:
\begin{equation}
\pi_{\bar\theta}^{S}(a_t\mid h_t,S_x)
=
\pi_{\bar\theta}(a_t\mid h_t,c(S_x)).
\end{equation}
Starting from a candidate state $v$, the teacher interacts with the
environment and generates at most $N_T$ continuations. A continuation is
accepted only if
\begin{equation}
R(\tau_v^T)\geq r_{\mathrm{succ}}.
\end{equation}
If no continuation succeeds, OVCSD moves to the nearest eligible ancestor and
retries. This backtracking procedure adaptively expands how much of the failed
student trajectory the teacher must replace.

The acceptance criterion depends only on the executed terminal outcome, not
on the teacher's local likelihood. Once a verified continuation is found, it
supervises the failed branches in $M(v)$; covered branches are not supervised
again by ancestor nodes.

\subsection{Alignment-Aware Comparative Learning}

A verified teacher continuation is not distilled uniformly. As established by
Observation~3, its relationship with a failed student branch changes at their
first aligned divergence. Before the divergence, their behavior is identical.
At the divergence, they take different actions from the same state and can be
directly compared. Afterward, their histories separate and no longer support
state-wise action comparison. OVCSD therefore uses two complementary
objectives.

\paragraph{Correcting the first divergent decision.}
For a verified teacher continuation from node $v$ and a supported failed
trajectory $i\in M(v)$, let $d_{v,i}$ denote their first action difference
while their states and preceding actions remain aligned:
\begin{equation}
\begin{aligned}
d_{v,i} = d_v + \min \Bigl\{ & j : a_{v,j}^{\mathrm{T}} \neq a_{i,d_v+j}^{0}, \\
& \operatorname{Align}\left(h_{v,j'}^{\mathrm{T}}, h_{i,d_v+j'}^{0}\right)=1, \; \forall j'\leq j \Bigr\}.
\end{aligned}
\end{equation}
Pairs whose alignment breaks before an action difference are discarded.
Student branches sharing the same divergence site $(v,d)$ form a local
comparison group. Let $m_{v,d}$ be the number of matched failed branches.
The verified teacher action receives positive advantage, while each failed
student alternative receives negative advantage:
\begin{equation}
A_{v,d}^{T}=\sqrt{m_{v,d}},
\qquad
A_{v,d,i}^{0}=-\frac{1}{\sqrt{m_{v,d}}}.
\end{equation}
This normalization balances the total positive and negative advantage mass at
each divergence site.

OVCSD activates only the divergent action tokens in the matched student rows
and appends one masked teacher-action row at the same state. Because the
teacher action was generated with privileged context, its likelihood ratio is
recomputed under the frozen skill-free policy before applying the standard
clipped GRPO update. All other rows retain their original group-relative
advantages.

\paragraph{Distilling the completion suffix.}
The local comparative update identifies the divergent teacher action but does not
specify how to complete the remaining task. OVCSD therefore distills the
teacher actions strictly after the divergence.

At each post-divergence teacher token, we compare the detached privileged
teacher distribution
$q_{t,m}=\operatorname{stopgrad}
[\pi_{\bar\theta}^{S}(\cdot\mid\widetilde h_{t,m}^{T},S_x)]$
with the skill-free student distribution
$p_{t,m}=\pi_{\theta}^{0}(\cdot\mid\widetilde h_{t,m}^{T})$.
The suffix objective is
\begin{equation}
\mathcal{L}_{\mathrm{suf}}
=
\frac{1}{|\mathcal D^{+}|}
\sum_{(v,d)\in\mathcal D^{+}}
\frac{1}{|\Omega_{v,d}|}
\sum_{(t,m)\in\Omega_{v,d}}
D_{\mathrm{KL}}\!\left(q_{t,m}\,\|\,p_{t,m}\right),
\end{equation}
where $\Omega_{v,d}$ contains action-token positions strictly after the
divergence. In practice, we compute the divergence over the teacher's top-$k$
support and aggregate the remaining probability mass; details are given in
supplementary material.

\subsection{Overall Objective and Inference}

The complete training objective is
\begin{equation}
\mathcal{L}_{\mathrm{OVCSD}}
=
\mathcal{L}_{\mathrm{GRPO}}^{\mathrm{aug}}
+
\lambda_{\mathrm{suf}}\mathcal{L}_{\mathrm{suf}},
\end{equation}
together with the standard regularization terms of the reinforcement-learning
backbone. The first term learns which action should replace a failed decision
at a shared state; the second transfers the behavior needed to complete the
task afterward.

All privileged operations are used only during training. At inference time,
OVCSD directly deploys the skill-free policy $\pi_\theta^0$ without skill
retrieval, prefix-tree construction, environment replay, or teacher
interaction.

\begin{table}[t]
\centering
\small
\renewcommand{\arraystretch}{1.08}
\setlength{\tabcolsep}{1.3pt}

\begin{tabular}{@{}l ccccccccc @{}}
\toprule
\multicolumn{1}{l}{} &
\multicolumn{7}{c}{\textbf{ALFWorld}} &
\multicolumn{2}{c}{\textbf{WebShop}} \\
\cmidrule(lr){2-8}\cmidrule(lr){9-10}
Method &
Pick & Look & Clean & Heat & Cool & Pick2 & Avg &
Score & Acc \\
\midrule

\multicolumn{10}{@{}>{\columncolor{gray!12}[0pt][\tabcolsep]}l}{\hspace{4pt}\textbf{Qwen3-1.7B}} \\
Vanilla & 25.0 & 22.2 & 3.1 & 0.0 & 21.4 & 4.2 & 12.5 & 46.5 & 4.7 \\
SkillCOT* & 10.5 & \second{50.0} & 16.1 & 0.0 & 0.0 & 5.0 & 9.4 & 23.0 & 2.3 \\
GRPO & 71.1 & 41.7 & 36.4 & 40.0 & 31.8 & 31.6 & 46.1 & 67.3 & 38.3 \\
OPSD & 26.3 & 33.3 & 9.1 & 0.0 & 4.5 & 5.3 & 14.1 & 47.4 & 9.3 \\
Skill-SD & 52.9 & 37.5 & 69.2 & \second{42.9} & \second{60.0} & \second{36.8} & 52.3 & \best{81.8} & 53.9 \\
SDAR & \second{73.5} & 25.0 & \second{76.9} & 33.3 & 40.0 & \second{36.8} & \second{53.9} & 76.8 & \second{58.6} \\
OVCSD & \best{96.9} & \best{83.3} & \best{91.7} & \best{75.0} & \best{71.4} & \best{74.8} & \best{83.6} & \second{77.9} & \best{62.5} \\
\midrule

\multicolumn{10}{@{}>{\columncolor{gray!12}[0pt][\tabcolsep]}l}{\hspace{4pt}\textbf{Qwen2.5-3B}} \\
Vanilla & 44.4 & 11.1 & 6.2 & 15.4 & 28.6 & 12.5 & 21.9 & 6.7 & 0.8 \\
SkillCOT* & 51.7 & \second{66.7} & 48.4 & 0.0 & 4.3 & 10.0 & 28.9 & 0.2 & 0.8 \\
GRPO & \second{91.2} & 62.5 & \second{96.2} & \second{61.9} & 65.0 & 47.4 & 75.0 & 79.8 & 63.3 \\
OPSD & 48.8 & 41.7 & 16.7 & 0.0 & 15.8 & 16.7 & 28.1 & 11.3 & 3.1 \\
Skill-SD & 88.2 & 50.0 & \second{96.2} & 52.4 & 65.0 & 57.9 & 73.4 & 75.9 & 64.0 \\
SDAR & \best{97.1} & 62.5 & \best{100.0} & \second{61.9} & \second{75.0} & \second{84.2} & \second{84.4} & \second{85.0} & \second{68.0} \\
OVCSD & 89.6 & \best{75.0} & 89.6 & \best{95.2} & \best{83.9} & \best{88.8} & \best{89.1} & \best{85.3} & \best{73.4} \\
\midrule

\multicolumn{10}{@{}>{\columncolor{gray!12}[0pt][\tabcolsep]}l}{\hspace{4pt}\textbf{Qwen2.5-7B}} \\
Vanilla & 36.1 & 22.2 & 3.1 & 0.0 & 0.0 & 0.0 & 12.5 & 5.9 & 1.6 \\
SkillCOT* & 51.7 & 50.0 & 32.3 & 5.3 & 4.3 & 0.0 & 23.4 & 1.7 & 0.8 \\
GRPO & 91.2 & \second{87.5} & 96.2 & 81.0 & 65.0 & 57.9 & 81.2 & 80.9 & 72.6 \\
OPSD & 50.0 & 60.0 & 22.7 & 21.4 & 17.6 & 9.5 & 32.8 & 4.5 & 2.3 \\
Skill-SD & 93.9 & \best{93.8} & 90.9 & \best{100.0} & \second{69.2} & 68.4 & 85.1 & 86.1 & 76.5 \\
SDAR & \second{94.7} & 75.0 & \best{100.0} & \second{86.7} & 68.2 & \second{78.9} & \second{85.9} & \second{89.4} & \second{82.8} \\
OVCSD & \best{98.2} & 83.3 & \second{97.9} & \best{100.0} & \best{82.1} & \best{94.4} & \best{92.2} & \best{91.2} & \best{84.4} \\
\bottomrule
\end{tabular}
\caption{Performance on ALFWorld and WebShop tasks. We report success rate (\%) on ALFWorld and score/accuracy (\%) on WebShop. WebShop Score is reported as the task score multiplied by 100. * denotes evaluation with retrieved skills at inference time. \best{Bold with yellow box} and \second{underlined with blue box} mark the best and second-best results within each base-model block. All methods except SkillCOT* use the skill-free deployment prompt.}
\label{tab:main-results}
\end{table}

\section{Experiments}
We investigate four questions.
\textbf{(Q1)} Does OVCSD improve skill-free performance across model scales?
\textbf{(Q2)} Is the gain bought with the extra environment interactio?
\textbf{(Q3)} Which design commitments carry the gain?
\textbf{(Q4)} What does teacher intervention change in the policy?
 
\subsection{Experimental Setup}
\paragraph{Benchmarks and metrics.}
We evaluate on ALFWorld~\cite{alfworld}, which contains multi-step household
tasks involving navigation, object manipulation, and state transformation, and
WebShop~\cite{webshop}, which requires an agent to search for and purchase
products satisfying multiple natural-language constraints. For ALFWorld, we
report success rates on the Pick, Look, Clean, Heat, Cool, and Pick2 categories,
together with the micro-averaged success rate over all evaluation tasks. For
WebShop, we report the official score multiplied by 100 and the success rate.

\paragraph{Models and baselines.}
We use Qwen3-1.7B, Qwen2.5-3B, and Qwen2.5-7B. Within each model configuration,
we use the same student rollout group size, reward function, and GRPO schedule. Vanilla denotes the supervised initialization before reinforcement learning, while
SkillCOT* augments it with retrieved skills at inference time. GRPO serves as
the standard skill-free reinforcement-learning baseline. We further compare
against three privileged self-distillation methods evaluated without skills at
inference: OPSD~\cite{opsd}, Skill-SD~\cite{skillsd}, and
SDAR~\cite{sdar}. OVCSD likewise uses skills only during training and deploys
the resulting policy without skill retrieval. Except for SkillCOT*, all reported
results are obtained under the skill-free prompt.

\subsection{Main Results (Q1)}
Table~\ref{tab:main-results} reports ALFWorld and WebShop. OVCSD achieves the
best micro-averaged ALFWorld success and the best WebShop success at all three
scales. Against GRPO it improves ALFWorld success by 37.5, 14.1, and 11.0
points at 1.7B, 3B, and 7B, and WebShop success by 24.2, 10.1, and 11.8 points;
WebShop score improves by 10.6, 5.5, and 10.3 points. It obtains the best
WebShop score at 3B and 7B and ranks second at 1.7B, and exceeds the strongest
skill-free baseline, SDAR, on ALFWorld by 29.7, 4.7, and 6.3 points. The gains
hold from 1.7B to 7B.
 
Gains concentrate on the compositional Heat, Cool, and Pick2 categories, where
OVCSD is best or tied at every scale; Pick2 improves over GRPO by 43.2, 41.4,
and 36.5 points. Where strong baselines are near saturation the picture is
mixed: OVCSD does not lead Pick or Clean at 3B, or Look at 7B. These results suggest that OVCSD provides its largest gains on tasks involving multiple dependent decisions.
 
\subsection{Interaction Budget (Q2)}
\label{sec:budget}

\begin{figure}[t]
\centering
\includegraphics[width=0.38\textwidth]{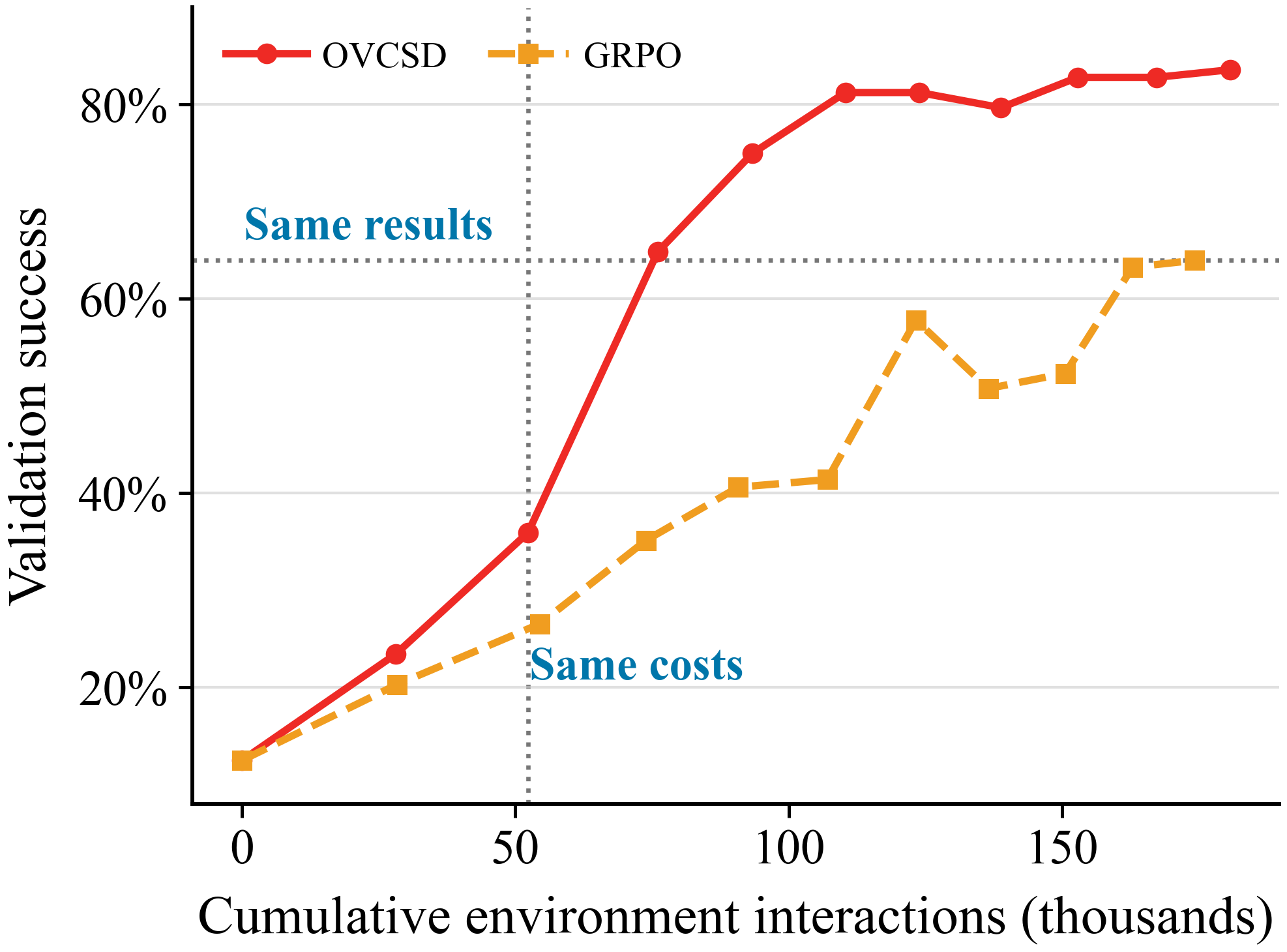}
\caption{Skill-free validation success against cumulative environment
interactions on ALFWorld (1.7B). At a matched budget of 123k actions OVCSD
reaches 81.3\% against GRPO's 57.8\%. The shaded band marks budgets where only
OVCSD has completed validation.}
\label{fig:budget}
\end{figure}

Prefix replay and teacher continuations both touch the environment, so a gain
measured per gradient step could merely reflect a larger sampling budget. We
therefore compare against cumulative environment actions.

\paragraph{Privileged interaction is a small fraction of the budget.}
Table~\ref{tab:cost} decomposes the budget over 100 updates. Replay plus teacher
continuations stay near 2\% of environment steps throughout, an effective group
size of 8.17 against a nominal 8. Raising the nominal group size from $K{=}8$ to
$K{=}9$ would instead add 12.5\% more rollout, over-matching the measured
overhead several times over; we therefore treat the budget-aligned curve as the
matched comparison rather than running that baseline. Within that overhead,
intervention fires on 49 of 100 updates and intercepts 70 all-failure groups,
consuming 92 teacher attempts of which only 52 complete the task: outcome
verification rejects 43.5\% of skill-conditioned continuations, and 74.3\% of
intercepted groups eventually yield a verified branch.

\begin{table}[t]
\centering
\small
\setlength{\tabcolsep}{6pt}
\renewcommand{\arraystretch}{1.1}
\begin{tabular}{@{}lrrrrr@{}}
\toprule
Update & Student & Replay & Teacher & Overhead & Eff.\ $K$ \\
\midrule
30  &  74{,}084 & 152 & 1{,}818 & 2.59\% & 8.21 \\
60  & 121{,}377 & 288 & 2{,}267 & 2.06\% & 8.17 \\
100 & 176{,}961 & 492 & 3{,}314 & 2.11\% & 8.17 \\
\bottomrule
\end{tabular}
\caption{Cumulative environment action steps by source on ALFWorld over 100
updates. \emph{Overhead} is replay plus teacher continuations as a share of the
total; \emph{Eff.\ $K$} rescales the nominal group size of 8 by that total.
Privileged interaction stays near 2\% throughout training.}
\label{tab:cost}
\end{table}

\paragraph{The gain is not bought with interaction.}
Figure~\ref{fig:budget} separates the two curves from the first validation point
onward, so OVCSD does not trade early cost for late benefit. At the matched
budget of 123k actions it leads by 23.5 points (81.3\% against 57.8\%), and
reaches GRPO's final success after 81.4k actions against GRPO's 174.3k, a
2.1$\times$ reduction, with privileged interaction near 2\% of the budget.

\subsection{Ablations (Q3)}
\label{sec:ablation}

The budget-aligned comparison in Q2 indicates that OVCSD's gain cannot be
explained solely by additional environment interaction. It leaves open whether
the gain needs \emph{this} mechanism: branching from replayed failure states and
accepting each continuation on its own terminal reward. A simpler design would
rerun the skill-conditioned teacher from the initial state and distill whatever
succeeds. We ablate the two commitments that separate OVCSD from that design,
and the two uses it makes of a verified branch.
Every variant starts from the same supervised checkpoint, trains for 50 updates
under an identical schedule, reward, and group size, and is evaluated skill-free
on the 128-task validation set.

\paragraph{How the branch is transferred.}
As shown in Table~\ref{tab:component}, neither signal alone recovers the
performance of full OVCSD: ablating Local-PG or Suffix-KL drops performance from
81.25\% to 66.41\% and 64.06\%, respectively. The remaining performance gap
supports that decision-level correction and trajectory-completion supervision
provide complementary learning signals.

\begin{table}[t]
\centering
\small
\renewcommand{\arraystretch}{1.15}
\begin{tabular}{@{} l @{\hspace{14pt}} c @{\hspace{10pt}} c @{\hspace{16pt}} c @{}}
\toprule
Method & Local $\hat{A}$ & Suffix KL & Success \\
\midrule
GRPO           &            &            & \refval{41.41} \\
Local-PG only  & \checkmark &            & \gain{66.41}{25.00} \\
Suffix-KL only &            & \checkmark & \gain{64.06}{22.65} \\
\midrule
\textbf{OVCSD} & \checkmark & \checkmark & \gain{81.25}{39.84} \\
\bottomrule
\end{tabular}
\caption{Objective ablation on skill-free ALFWorld validation (\%). Local
$\hat{A}$ is the localized advantage at the divergence state, Suffix KL the
top-$k$ distillation on the completion suffix. Subscripts are gains over GRPO.}
\label{tab:component}
\end{table}

\paragraph{How the branch is acquired.}
Table~\ref{tab:intervention} compares intervention anchoring and outcome
verification. Although final success is similar, deepest-first anchoring
preserves more student progress and reduces the cost per verified branch from
33.00 to 25.60 actions: replay lengthens from 1.73 to 3.56 actions per attempt
while teacher execution shortens from 18.90 to 13.75, and the latter dominates.
Without outcome verification, 43.9\% of the branches accepted for training fail
to complete the task. OVCSD filters these invalid targets and produces more
verified branches (48 versus 37) at a lower cost per verified branch (25.60
versus 30.57).

Deepest-first does not simply take the deepest state on offer.
Table~\ref{tab:depth} contrasts the deepest state-aligned candidate available in
an intervened group with the state actually launched from: mass shifts toward
shallower depths by 0.88 actions on ALFWorld and 0.18 on WebShop, and on
ALFWorld no group is root-only yet 7.4\% of attempts start at the root, which
can only follow a backoff. WebShop admits no candidate beyond the root in 42.1\%
of groups against 0\% on ALFWorld, so the intervention path fires more often per
ALFWorld update (Observation~1).

\begin{table}[t]
\centering
\footnotesize
\setlength{\tabcolsep}{4pt}
\begin{tabular}{@{}lrrrrr@{}}
\toprule
Strategy & Success & Attempts & Verified & Depth & Cost/ver. \\
\midrule
Random anchor   & 79.69 & 40 & 25 & 1.73 & 33.00 \\
No verification & 80.47 & 66 & 37 & 2.74 & 30.57 \\
\textbf{OVCSD}  & \textbf{81.25} & 71 & 48 & 3.56 & \textbf{25.60} \\
\bottomrule
\end{tabular}
\caption{Intervention ablation over 50 updates. \emph{Attempts} are teacher
continuations executed, \emph{Verified} those that complete the task;
\emph{Depth} is mean replay depth and \emph{Cost/ver.} the replay plus teacher
actions spent per verified branch. Attempt counts are an outcome of each run,
not a matched input; the controlled comparison is cost per verified branch.}
\label{tab:intervention}
\end{table}

\begin{table}[t]
\centering
\small
\setlength{\tabcolsep}{5pt}
\renewcommand{\arraystretch}{1.05}
\begin{tabular}{@{}lcccc@{}}
\toprule
& \multicolumn{2}{c}{ALFWorld} & \multicolumn{2}{c}{WebShop} \\
\cmidrule(lr){2-3}\cmidrule(lr){4-5}
Depth & Avail. & Used & Avail. & Used \\
\midrule
0    & --   &  7.4 & 42.1 & 33.2 \\
1    & --   &  7.4 &  4.7 & 15.1 \\
2    &  5.3 & 11.1 & 12.8 & 18.1 \\
3    & 21.1 & 18.5 &  6.8 & 16.2 \\
4    & 26.3 & 22.2 & 30.2 & 16.0 \\
5    & 31.6 & 22.2 &  3.4 &  1.4 \\
6+   & 15.9 & 11.1 & --   & --   \\
\midrule
Mean & 4.58 & 3.70 & 1.89 & 1.71 \\
\bottomrule
\end{tabular}
\caption{Distribution (\%) of intervention launch depth. \emph{Avail.} is the
deepest state-aligned candidate available in an intervened group, \emph{Used}
the state the teacher was launched from, \emph{Mean} in actions. Statistics pool
all intervened groups in the archived main runs. }
\label{tab:depth}
\end{table}
\subsection{What Intervention Changes (Q4)}
\label{sec:mechanism}
We measure the policy change with forward-only diagnostics over archived
states, using no replay and no resampling.
 
\paragraph{The policy moves toward teacher decisions on verified branches.}
Table~\ref{tab:internalization} examines whether training increases the probability of the teacher action taken at each verified branch's aligned divergence. Relative to the shared SFT initialization, teacher-action log-probability increases by $0.342$ under OVCSD, compared with $0.060$ under GRPO, yielding a between-method difference of $+0.282$ $[0.17, 0.41]$. The teacher-action probability increases at 91.0\% of aligned divergence states under OVCSD, compared with 67.9\% under GRPO. The shift is directional rather than a uniform increase in likelihood at these states: relative to GRPO, OVCSD increases the teacher-versus-student preference by $+0.309$ $[0.08, 0.54]$ and the teacher-versus-control preference by $+0.235$ $[0.01, 0.47]$. These diagnostics indicate that OVCSD induces a localized preference for the teacher action observed on the successful recovery branch.
 
\begin{table}[t]
\centering
\small
\setlength{\tabcolsep}{4pt}
\renewcommand{\arraystretch}{1.1}
\begin{tabular}{@{}lrrrc@{}}
\toprule
Scored action & GRPO & OVCSD &$\Delta$ & 95\% CI \\
\midrule
Verified teacher    & $+0.060$ & $+0.342$ & $+0.282$ & $[0.17, 0.41]$ \\
Failed student      & $-0.054$ & $-0.081$ & $-0.027$ & $[-0.24, 0.17]$ \\
Prefix control      & $-0.208$ & $-0.162$ & $+0.046$ & $[-0.13, 0.23]$ \\
\midrule
States increased (\%) & $67.9$ & $91.0$ & $+23.1$ & $[12.8, 34.6]$ \\
\bottomrule
\end{tabular}
\caption{Change in teacher-forced action-span log-probability relative to the
shared SFT initialization.}
\label{tab:internalization}
\end{table}
 
\paragraph{Policy change concentrates near the divergence.}
Over training progress, OVCSD places $1.56\times$ $[1.26, 1.89]$ more KL mass near the aligned divergence than at positions elsewhere in the trajectory. The
interval excludes $1.0$, so the concentration is a reliable effect; for GRPO the same quantity is $1.14\times$ $[0.90, 1.40]$, which is indistinguishable from no concentration at all. The paired difference of $+0.42$ $[0.07, 0.77]$ confirms the gap between methods, and the KL peak falls within $d \pm 1$ for 15.7\% of pairs under OVCSD against 6.7\% under GRPO, a $2.3\times$ increase. Policy change is thus concentrated where the method intervenes, while remaining spread over the suffix that OVCSD also distills.

\subsection{Case Study}
\begin{figure}[t]
\centering
\includegraphics[width=0.5\textwidth]{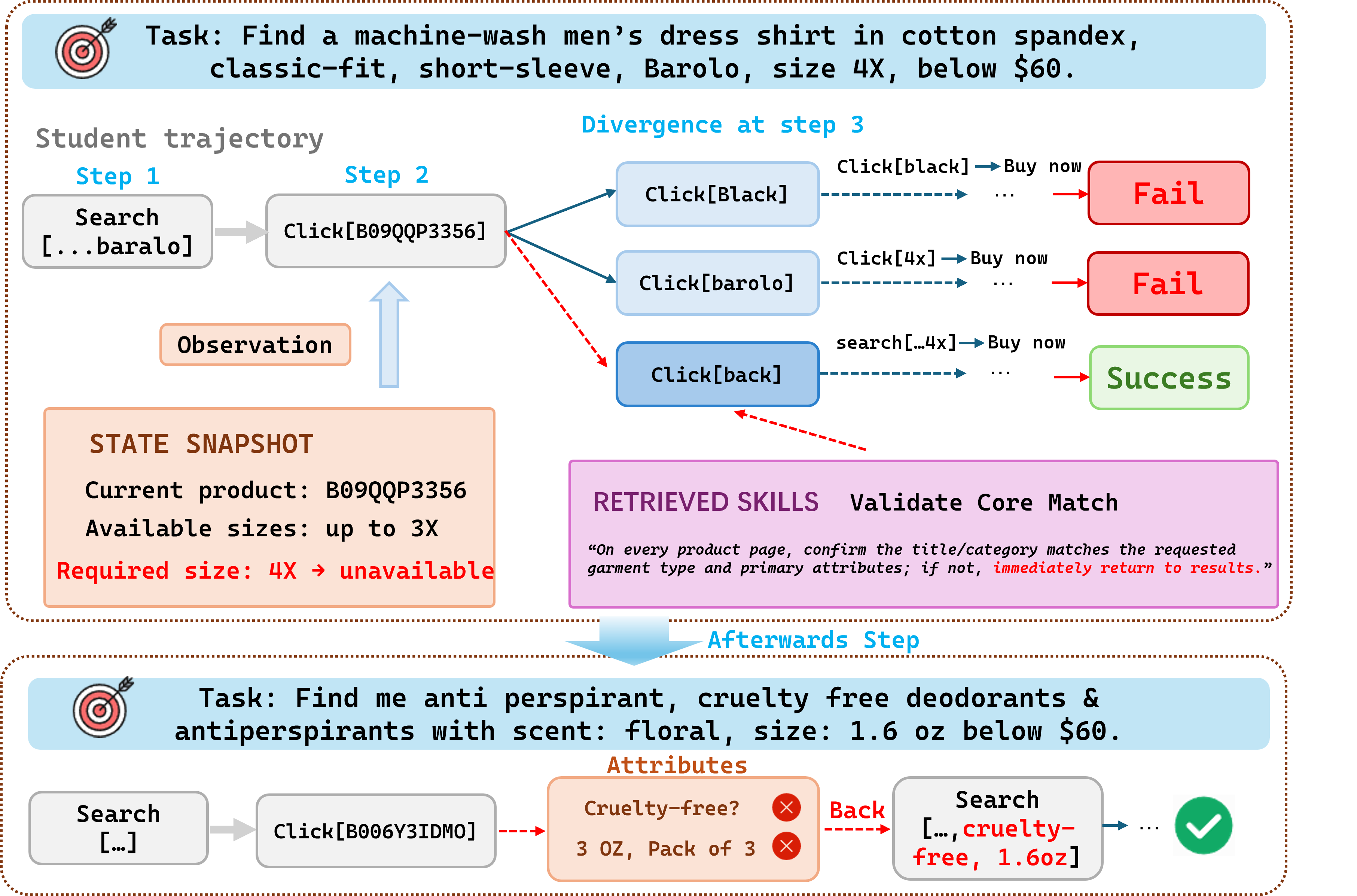}
\caption{Case study.}
\label{fig:case-study}
\end{figure}

Figure~\ref{fig:case-study} shows a locally verified correction becoming
reusable behavior. In the shirt task, student trajectories reach a product page
whose size inventory stops at 3X although the task requires 4X; continuing to
select options fails, whereas the skill-conditioned teacher applies Validate
Core Match, returns to search, and succeeds after adding the missing
constraint. In another task the skill-free student meets an Old Spice
three-pack with a fresh scent while the target requires a cruelty-free floral
1.6-ounce single pack; it abandons the mismatched candidate, repairs the query
with the omitted attributes, and completes the task. The two tasks differ in
product, target attribute, and omitted constraint, yet the same pattern appears in both. 
 
\section{Conclusion}
We presented Outcome-Verified Comparative Self-Distillation (OVCSD), an on-policy framework for internalizing privileged capabilities into skill-free agents. OVCSD adaptively constructs successful teacher continuations from student-reached states, verifies them through environment outcomes, and exploits their relationship with failed student rollouts through divergence-focused comparison and suffix distillation. Experiments on ALFWorld and WebShop demonstrate that OVCSD consistently improves long-horizon task performance over reinforcement-learning and self-distillation baselines, showing that outcome-verified and state-aligned teacher supervision provides an effective approach to capability internalization. 
\section{AI Discourse}
We used a large language model only to assist with prose polishing and plotting.

\appendix

\bibliography{aaai2027}


\end{document}